\documentclass[10pt,twocolumn,letterpaper]{article}

\usepackage{btas}
\usepackage{times}
\usepackage{epsfig}
\usepackage{graphicx}
\usepackage{amsmath}
\usepackage{amssymb}
\usepackage{subcaption}
\usepackage[norule,symbol,perpage]{footmisc}

\btasfinalcopy 

\ifbtasfinal\fi
\begin{document}


\title{Deep Learning-Based Feature Extraction in Iris Recognition: Use Existing Models, Fine-tune or Train From Scratch?}

\author{Aidan Boyd, Adam Czajka, Kevin Bowyer\\
University of Notre Dame\\
Notre Dame, Indiana\\
{\tt\small \{aboyd3, aczajka, kwb\}@nd.edu}
}


\maketitle
\thispagestyle{empty}

\begin{figure*}[h!]
\includegraphics[scale = 0.53]{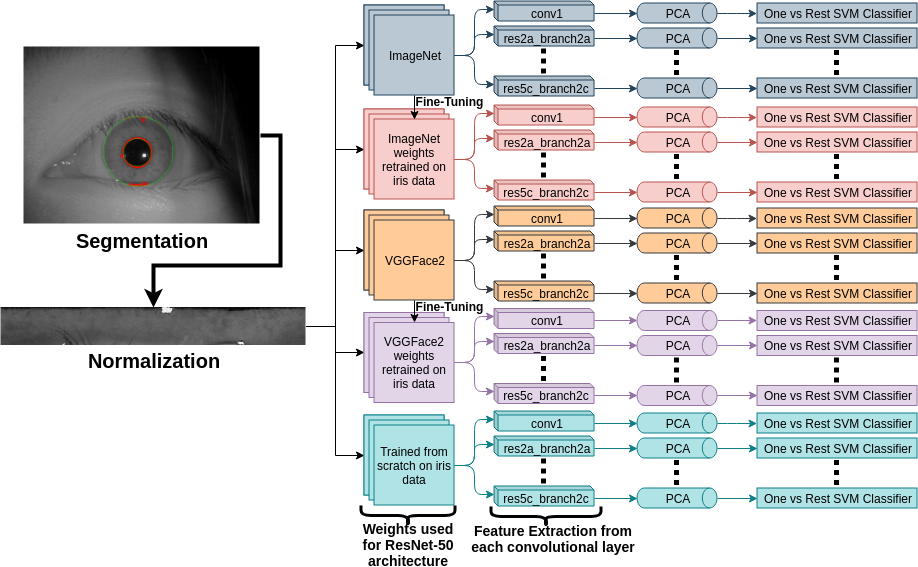}
\caption{Conceptual overview of experiments in this work. Left: Iris images from our in-house corpus (more than 370,000 iris images) and CASIA-Iris-Thousand are segmented with the OSIRIS to create network training and classification training/testing datasets, respectively.  Middle: ResNet-50 trained on ImageNet, ResNet-50 trained on VGGFace, both of these fine-tuned on 360K+ iris training images, and ResNet-50 trained from scratch on the same 360K+ training iris image set are used to generate feature vectors from each of the convolutional layers. Right: Classification on the CASIA-Iris-Thousand iris image set that is subject-disjoint and cross-sensor relative to the in-house 370K+ iris image set used in CNN training is used to compute accuracy for comparison of approaches.}
\label{fig: pipeline}
\end{figure*}

\begin{abstract}
Modern deep learning techniques can be employed to generate effective feature extractors for the task of iris recognition. The question arises: should we train such structures from scratch on a relatively large iris image dataset, or it is better to fine-tune the existing models to adapt them to a new domain? In this work we explore five different sets of weights for the popular ResNet-50 architecture to find out whether iris-specific feature extractors perform better than models trained for non-iris tasks. Features are extracted from each convolutional layer and the classification accuracy achieved by a Support Vector Machine is measured on a dataset that is disjoint from the samples used in training of the ResNet-50 model. We show that the optimal training strategy is to fine-tune an off-the-shelf set of weights to the iris recognition domain. This approach results in greater accuracy than both off-the-shelf weights and a model trained from scratch. The winning, fine-tuned approach also shows an increase in performance when compared to previous work, in which only off-the-shelf (not fine-tuned) models were used in iris feature extraction. We make the best-performing ResNet-50 model, fine-tuned with more than 360,000 iris images, publicly available along with this paper.
\end{abstract}

\section{Introduction}

The task of developing reliable feature extractors for iris recognition is still an open research problem. 
Iris recognition has gained a position as one of the fastest and most accurate biometric recognition methods, deployed in several large-scale national ID \cite{AADHAAR} and border control \cite{NEXUS} programs. The approach of translating the output of Gabor filtering into a binary code, proposed more than 25 years ago \cite{Daugman_PAMI_1993}, dominates current commercial implementations. Using present trends in machine learning and explaining this approach in the language of convolutional neural networks (CNN), the Daugman's method of iris code generation could be visualized as a single convolutional layer with neurons having {\it hardlim} \cite{hardlim-matlab} activation functions. Although this structure seems to be simple, it is not necessarily a trivial task to find a set of kernels implemented by this single convolutional layer to extract salient iris features. This task does not become significantly simpler even if we narrow ourselves to Gabor wavelets.

Convolutional neural networks, recently very successful in solving various computer vision tasks, have been also shown to serve as good iris feature extractors \cite{Nguyen_Access_2018}. These structures are certainly more complex than Daugman's approach, but the fact that there is no need to search for optimal convolutional kernels, and thus the off-the-shelf architectures can be directly used in iris recognition, is appealing. However, intuitively the domain-specific image processing methods should perform better than general-purpose ones, as it has been shown also for iris recognition \cite{Czajka_WACV_2019}. In this paper we present experiments and answer the following two questions:

\begin{enumerate}
    \item[Q1.] Which models perform better in iris recognition: off-the-shelf, \ie, not requiring training with iris data, or trained with iris images?
    \item[Q2.] If it is better to use trained models, what training strategy is better: train from scratch on relatively large set of iris images, fine-tune a model designed for a general image recognition task, or fine-tune the model used for face recognition?
\end{enumerate}

For that purpose we use the ResNet-50 model \cite{he2016deep} and a set of more than 360,000 iris training images. An additional set of 20,000 subject-disjoint iris images are used for classifier training and testing. The fine-tuned models, which achieved higher accuracy than off-the-shelf networks in our experiments, are made available along with the paper.

\section{Related Work}

Convolutional Neural Networks have been employed to achieve state-of-the-art iris recognition performance. Liu \etal proposed DeepIris \cite{liu2016deepiris}, the first deep learning method for heterogeneous iris verification. The authors proposed a nine layer architecture including a single convolutional layer, two pooling layers, two normalization layers, two local layers and one fully connected layer. Experimental results validate the effectiveness of applying CNNs to iris recognition by attaining promising results for both cross-resolution and cross-sensor iris validation. Gangwar and Joshi \cite{gangwar2016deepirisnet} later proposed two deeper architectures for iris recognition. These two networks exhibited superior performance on the ND-IRIS-0405 \cite{bowyer2016nd} and ND-CrossSensor-Iris-2013 \cite{nddatasets} datasets. Proen\c{c}a and Neves \cite{proencca2017irina} reinforce the capabilities of neural networks by showing that their proposed model achieved state-of-the-art performance for recognition for good quality data while also being robust against segmentation errors and large changes in pupil sizes. 

Convolutional Neural Networks have also been shown to perform as effective feature extractors \cite{menotti2015deep,chen2016deep,mahmood2017resfeats}. In a work by Nguyen \etal \cite{Nguyen_Access_2018}, off-the-shelf weights are explored as feature extractors for the task of iris recognition. Five state of the art network architectures are examined and features are extracted layers at various stages of the network. Promising results are reported even though the off-the-shelf weights utilized were not trained for the task of iris recognition. {\bf Our paper is different} in a way that in their paper the results from the five tested architectures come from off-the-shelf weights. In our paper, we determine whether the fine-tuning of weight parameters increases performance.

Minaee \etal \cite{minaee2016experimental} also explored the use of deep convolutional features for iris recognition. In this work, the authors extract features from each layer of the VGG-Net \cite{simonyan2014very}. It is shown that these features result in high classification accuracy. {\bf Our paper is different} in a way that in their work, different weight configurations are not explored, instead they use the ImageNet weights to extract features. In our work, features are extracted in a similar way, however, instead of investigating which layer is most performant, we are exploring what is the best way to train the network to achieve the best results.

In a paper by Zanlorensi \etal \cite{zanlorensi2018impact}, fine-tuned face weights are used in both the ResNet-50 and VGG models to extract weights from the last layer of the network on iris data for the task of iris data augmentation and segmentation. They show that the use of transfer learning leads to the generation of good feature extractors. {\bf Our paper is different} in a way that in their work features are only extracted from the last layer before the classification layer of the architectures. In our work, features are extracted from each of the convolutional layers in the network and we see that the best performing layers are those from the middle of the network.

Menon and Mukherjee \cite{menon2018iris} also proposed a method of feature extraction using deep convolutional networks. In their work, they use fine-tuned models starting from \mbox{ImageNet} weights to extract features for the purpose of iris recognition. Features are extracted from the last layer before the classification layer and passed to two single layer perceptrons. The input to their proposed method is two iris images and the output is whether they are the same person or not. {\bf Our paper is different} in a way that we make use of a one-versus-rest SVM in which we pass in a single image and it outputs which class it belongs to.

\section{Methodology}

This section describes the experimental setup for this work. The weights of all trained networks as well as random seeds have been made available \cite{aboyd-github} such that tests can be reproduced.

\subsection{Databases}
The dataset used to train the network is a set of in-house iris data collected by the University of Notre Dame. This set consists of 2000 classes of irises, totalling of 373,629 full iris images. All images in this set are live irises without contact lenses. Images in this set were acquired using LG 2200, LG 4000 and IrisGuard AD100 sensors.

The dataset that was used for testing and classification was the CASIA-Iris-Thousand database \cite{casia-database}. This database contains 20,000 images from 1000 subjects, collected using the IKEMB-100 camera from IrisKing. Both left and right iris images were acquired meaning there are 2000 total classes in this database. 

To simplify the explanation of the different data subsets, labels have been assigned. The subset of our in-house data used to train the networks is labelled the {\it network training set}. The subset that will be used to train the classifier will be known as the {\it classification training set} and the remaining samples used to test the classifier will be known as the {\it classification test} set. The {\it classification training} and {\it classification test} sets are both independent splits of the CASIA-Iris-Thousand database. The classification training and test databases are both subject-disjoint and cross-sensor in comparison to the network training set.

\subsection{Segmentation}

The tool used to segment all iris images in this work is OSIRIS \cite{othman2016osiris}. OSIRIS locates the pupil and iris boundaries and generates normalized iris images of size $64\times512$. When segmenting the network training set, if the segmentation failed we excluded that sample entirely from the subset. Out of the 373,629 full iris images in the network training set, there were 10,117 failures (about 3 percent), meaning the final network training set is of size 363,512 normalized iris images. The reason for this data curation is to use valid training samples and let the network learn iris-related features.

When segmenting and normalizing the CASIA-Iris-Thousand database, there was only 27 failures, corresponding to less than 0.2\% error. For simplicity, failed samples were eliminated from the dataset meaning that the combined size of the classification training and classification testing set was 19,973 images from 2000 classes. One possible reason for the difference in performance between the network training set and the CASIA-Iris-Thousand database is that the OSIRIS tool was developed and tested using the CASIA-Iris-Thousand database.

The normalized iris images used in network training and classification are by default grayscale images. It was required for the ResNet architecture that these be converted to RGB. This was done by copying all pixel values from the original single channel across all three channels. 

\subsection{Network Architecture}
The chosen network architecture for this work is a deep convolutional neural network model based on the Residual Network architecture with 53 convolutional layers (ResNet-50) \cite{he2016deep}. 

ResNet-50 is a fully convolutional architecture. All weights in a convolutional layer are shared between kernels on each pixel of the image, meaning input image dimensions do not have an effect on the operation of the network. Only dense layers, located at the end of the network, depend on the number of classes, and since we do not use the classification layers of the off-the-shelf networks, it is acceptable to use any input size greater than $32\times32$ pixels, specified in the Keras ResNet-50 documentation \cite{keras-resnet}. This is important as the input to each of the networks in this work is the $64\times512\times3$ normalized iris image. Although the images used to train the off-the-shelf network were the default ResNet dimensions of $224\times224\times3$, these weights are still applicable to images of different sizes.

\subsection{Network Training}

In this work we examine five different sets of weights for the ResNet-50 architecture. Three of these are trained or fine-tuned using iris images, and the other two are off-the-shelf weights obtained from training on the ImageNet \cite{deng2009imagenet} and \mbox{VGGFace2} \cite{cao2018vggface2} datasets. The first trained network is initialized using random weights. We denote this as being trained from scratch. The second is when the training is initialized on \mbox{ImageNet} weights and then the weight parameters are tweaked to be domain specific to iris recognition. The last trained network is initialized using \mbox{VGGFace2} weights and the parameters are tweaked as with the \mbox{ImageNet} network to be domain specific to iris recognition. The off-the-shelf weights are the default \mbox{ImageNet} weights from the Keras ResNet-50 implementation\cite{keras-resnet} and the set of weights obtained from training on the \mbox{VGGFace2} dataset using the keras\_vggface package\cite{keras-vggface-package}. The two off-the-shelf weight sets are used as a comparison to determine whether the parameter fine-tuning process yields better feature extractors.

For network training, the final classification layer of the architecture is removed and replaced with a custom dense layer due to the increased number of 2000 iris classes from the network training set that are being classified. A global average pooling layer is placed before this final dense layer to transform the features to a vector of size 2048. The feature vector of size 2048 is created as this is the number of channels in the output of the previous layer.

\subsection{Feature Extraction}

As the networks are not trained for the classification of the CASIA-Iris-Thousand database, we cannot use these networks directly as classifiers. Instead, features are extracted from layers of the network in the hope that they are generalized to the task of iris recognition. In this work, features are extracted from the output of each of the 53 individual convolutional layers in the network. These features are in the form of a vector ranging from size 16,384 to size 524,288, depending on the convolutional layer. These vectors will be referred to as the {\it feature vectors}.
In order to make sure all features are on the same scale, Min-Max scaling is performed independently on each feature between a range of 0 and 1. This scaling preserves inter-feature variance while making sure that larger scaled features don't dominate the feature selection even though they may not necessarily be the best features for classification.

\subsection{Feature Space Dimensionality Reduction}

Because the feature vectors are of such a large scale, we reduce the dimensionality of the feature space prior to classification. For each layer, Principal Component Analysis (PCA) is carried out, and we project all features onto a new subspace having 2000 dimensions. From a classification standpoint, we want to limit the features to those that are most important while not using too many and therefore over-fitting to the data. Through experimentation, it was found that most feature vectors were reduced to within the 1000-2000 feature range after PCA, and 2000 dimensions was selected as a good number of features for the final experiments. The Singular Value Decomposition (SVD) solver that was used for PCA was ``randomized" as proposed in \cite{halko2011finding}. This was selected as it was shown to run faster than the default solver. Once the feature vector size was reduced to 2000, further reduction was made by selecting the number of features that correspond to 90\% of the feature variance. In some cases this did not result in any reduction from the 2000 features. PCA is employed mainly due the fact that an SVM is used as a classifier, which does not perform well with large dimensionality.

\subsection{Classification}

A one-versus-rest Support Vector Machine (SVM) is implemented with a linear kernel for classification. The classification training set is used to train these SVMs. Once the models have been created, these are tested using the classification testing set. The classification training set is 70\% of the CASIA-Iris-Thousand database and the classification testing subset is the remaining 30\%. The train/test split is stratified such that if there are 10 images for each class, seven will be used in training and the remaining three will be used for testing. This prevents scenarios where all samples from one class are in either the training or test set and therefore it is impossible to correctly classify these samples. A unique one-versus-rest classifier is created for each layer and the accuracy reported is how many correct classifications were made in the test set over the total number of samples in the classification test set. Linear kernels were selected as it was found that these performed best and in the least time.

\begin{figure*}[h!]
\includegraphics[scale = 0.37]{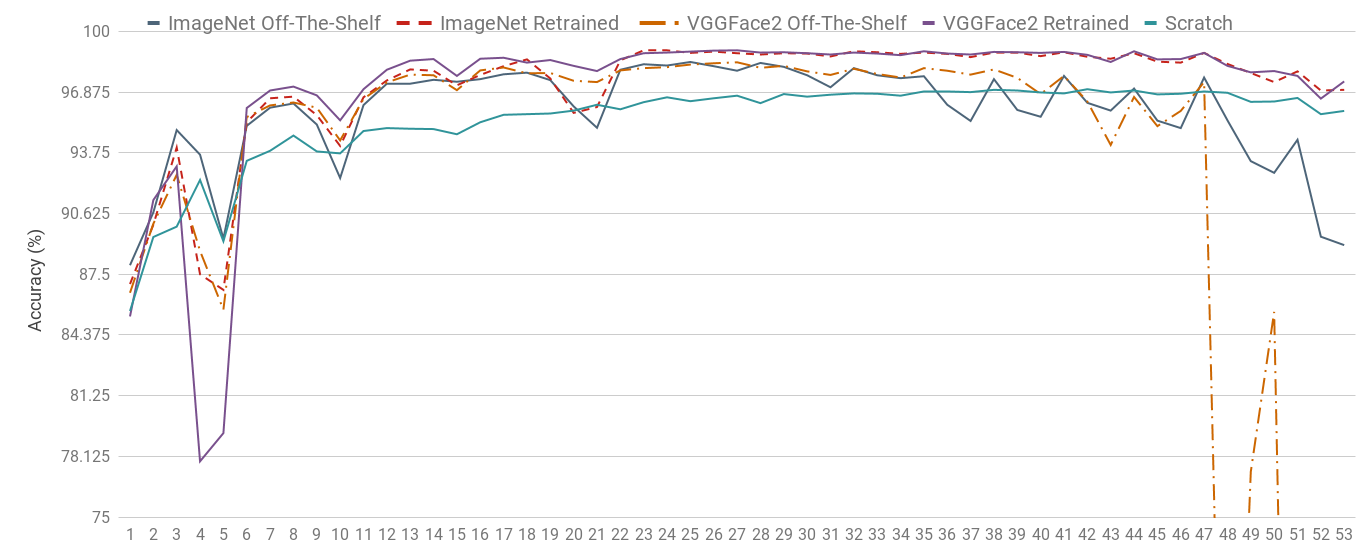}
\caption{This plot shows the classification accuracy for each convolutional layer of the five networks tested on the CASIA-Iris-Thousand dataset. The x-axis is the convolutional layer number. Out of frame: results of VGGFace2 off-the-shelf for layers 48, 51, 52 and 53 which were 47.4\%, 25.75\%, 39.87\% and 53.81\% respectively.}
\label{fig:results}
\end{figure*}

\section{Evaluation}

Figure \ref{fig:results} details the results obtained through experimentation. The x-axis of this graph is the number of the convolutional layer that the result was obtained from, \ie, layer 1 corresponds to the first convoltuional layer in the architecture after the input layer, and layer 53 is the final convolutional layer before the dense layer at the end. The names for these layers differ between the \mbox{ImageNet} networks and the \mbox{VGGFace2} networks. To find out the name for a layer number the list mapping layer numbers to names for both \mbox{ImageNet} and \mbox{VGGFace2} networks can be found in the repository \cite{aboyd-github}. The layer names for the trained from scratch network is the same as the \mbox{ImageNet} names. The y-axis is the classification accuracy, meaning how many correct classifications the SVM made over all total classifications. All classifications were made on a single random 70\%/30\% split of the CASIA-Iris-Thousand database into the classification training and classification testing sets. It was found that running these experiments on more than one split was infeasible due to the time required to run each. Analysis will now be done on all networks.

\subsection{Network Trained from Scratch}

As stated before, the training process for this network involved random weight initialization and then all weights are fine-tuned based on the network training set. It is evident from the Figure \ref{fig:results} that this network is the worst performing network, with most of the reported accuracy falling beneath the other four networks. Towards the later half of the network, however, we see these results stabilize and begin to perform consistently better than the off-the-shelf networks. This trained-from-scratch network performs worse than the two fine-tuned networks, as evident from Figure \ref{fig:results}. The reason for this performance may be due to the size of the network training set. This set contains 363,512 images from 2000 classes, which is very minimal in comparison to the quantity of data used to train the off-the-shelf networks. One interesting thing to note is the high number of layers that are achieving similar accuracy, namely in the second half of the network. It can be deduced that, even though the feature vectors for these layers are variant in size, they are describing features that result in similar classification accuracy. 

\subsection{Off-the-Shelf Networks}

The selected off-the-shelf configurations consisted of the weights used to classify the ImageNet database \cite{deng2009imagenet} and the weights used to classify the \mbox{VGGFace2} database \cite{cao2018vggface2}. The \mbox{ImageNet} weights used were the ``imagenet" weights for the Keras Implementation of ResNet-50 \cite{keras-resnet} and the \mbox{VGGFace2} weights were attained using the default implementation of ResNet-50 from the keras\_vggface package \cite{keras-vggface-package}. The results for these networks outlines the similarities between these weight sets at many of the layers. We see that in most cases in the first two thirds of the network, the \mbox{VGGFace2} off-the-shelf network performs slightly better than the \mbox{ImageNet} off-the-shelf. However, in the last 6 layers of the \mbox{VGGFace2} off-the-shelf architecture we see a drastic decrease in performance. In this same 6 layers, the \mbox{ImageNet} network performance also drops but not as extremely as VGGFace2. After some investigation, it was found that in these final layers of the VGGFace2 network that the PCA feature selection reduced the dimensionality to less than 100 features. It seems that the selected features had the highest variance but these did not contribute well to classification. In the layers that performed best, \ie, in the middle of the network, the feature vector size was reduced to between 500-2000. The last 6 layers (layers 47 to 53) in both of the off-the-shelf networks present lower and more variant results and as such can be seen as being the least useful as feature extractors. As none of the best results come from the last 6 layers in any of the architectures, these poor results in the \mbox{VGGFace2} off-the-shelf do not alter this work as we focus on only the best performing layers for each network. Layers in the middle of the architecture perform better and more stable. This PCA reduction is also the cause of the drop in accuracy seen by all networks in layers 4 and 5. Layers 4 and 5 must offer some features that are not useful for classification, and because this drop happens at the same layers in all 5 networks points to the possibility that it is an inherent feature of the ResNet architecture. 

At the early stages of the network the classification accuracy of both off-the-shelf networks is higher than that of the network trained from scratch, even though no iris domain information was used in the training of these networks. This outlines the generalization capabilities of these networks as feature extractors. This is an interesting result as it may outline the importance of the size of the network training set. Both of these networks were trained on datasets of much larger scale than our network training set. Both of these datasets used to generate the off-the-shelf weights had high heterogeneity present during training. The \mbox{ImageNet} weights are trained to classify thousands of classes of various images of largely variant subjects whereas the \mbox{VGGFace2} weights are trained to classify 9131 classes of faces. Although the accuracy of the off-the-shelf networks was not as high as the fine-tuned networks, they are still useful for iris recognition as multiple layers from both off-the-shelf networks obtained a classification accuracy of over 97.5\%.


\subsection{Fine-tuned Networks}

From looking at Figure \ref{fig:results}, it is clear that the fine-tuned networks are the highest performing. Both the fine-tuned \mbox{ImageNet} and fine-tuned \mbox{VGGFace2} weights perform similarly in many of the layers in the network. As with the network trained from scratch, the second half accuracy is stable. Fine-tuning the parameters from the \mbox{ImageNet} and \mbox{VGGFace2} weights to the iris network training set results in superior performance. One observation to be made here is that if training is done on a large heterogeneous dataset, this can be fine-tuned to a specific domain through weight retraining and achieve better results over training directly on the domain specific data. 

The results from these fine-tuned networks outline the effectiveness of this network as a feature extractor for iris recognition. These networks were fine-tuned using the network training set which is independent subject-disjoint and cross-sensor iris data to that it was tested on, the CASIA-Iris-Thousand database \cite{casia-database}, and classification accuracy as high as 99\% is reported for both the fine-tuned \mbox{ImageNet} and \mbox{VGGFace2} networks. It is evident that the network has learned efficient features that can be generalized to all iris data for recognition purposes. These results also display the benefits of transfer learning. Feature extractors from one domain can be effectively transferred to another domain through a process of fine-tuning.

\begin{figure*}[t]
  \begin{subfigure}[b]{1\textwidth}
      \begin{subfigure}[b]{0.327\textwidth}
          \centering
          \includegraphics[width=1\columnwidth]{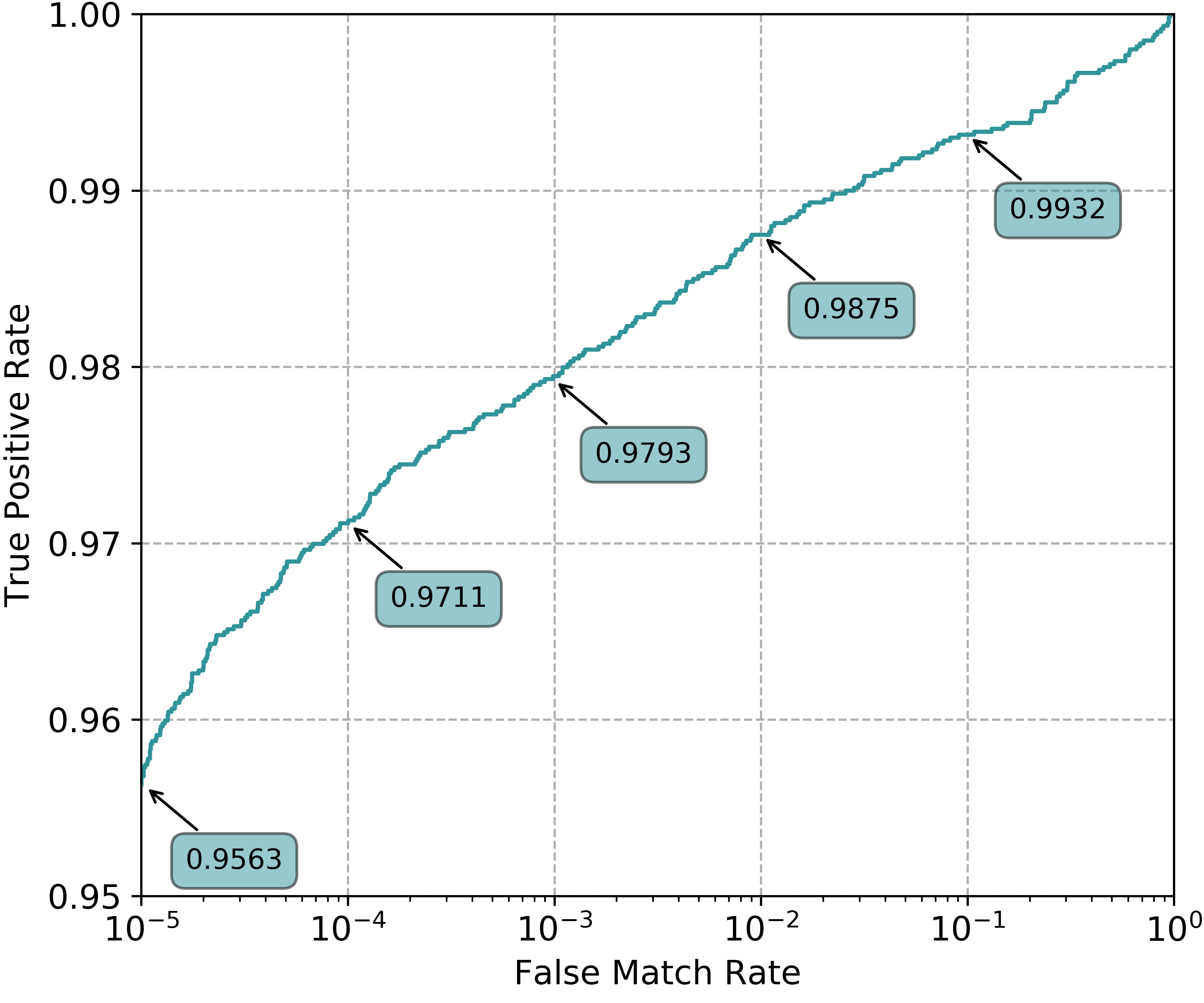}
          \caption{Trained From Scratch (Layer 42)}
      \end{subfigure}
      \hfill
      \begin{subfigure}[b]{0.327\textwidth}
          \centering
          \includegraphics[width=1\columnwidth]{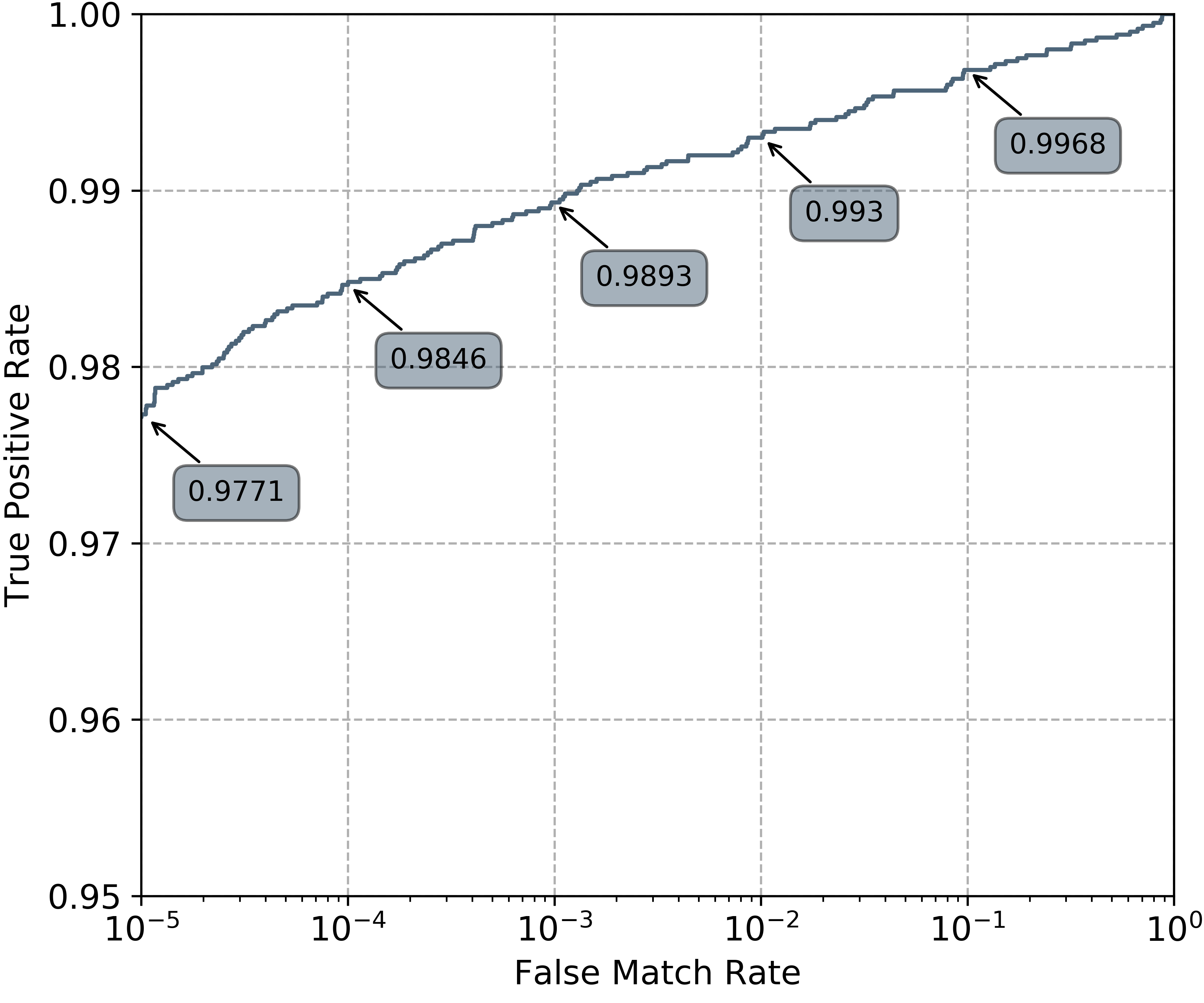}
          \caption{ImageNet Off-The-Shelf (Layer 25)}
      \end{subfigure}
      \hfill
      \begin{subfigure}[b]{0.327\textwidth}
          \centering
          \includegraphics[width=1\columnwidth]{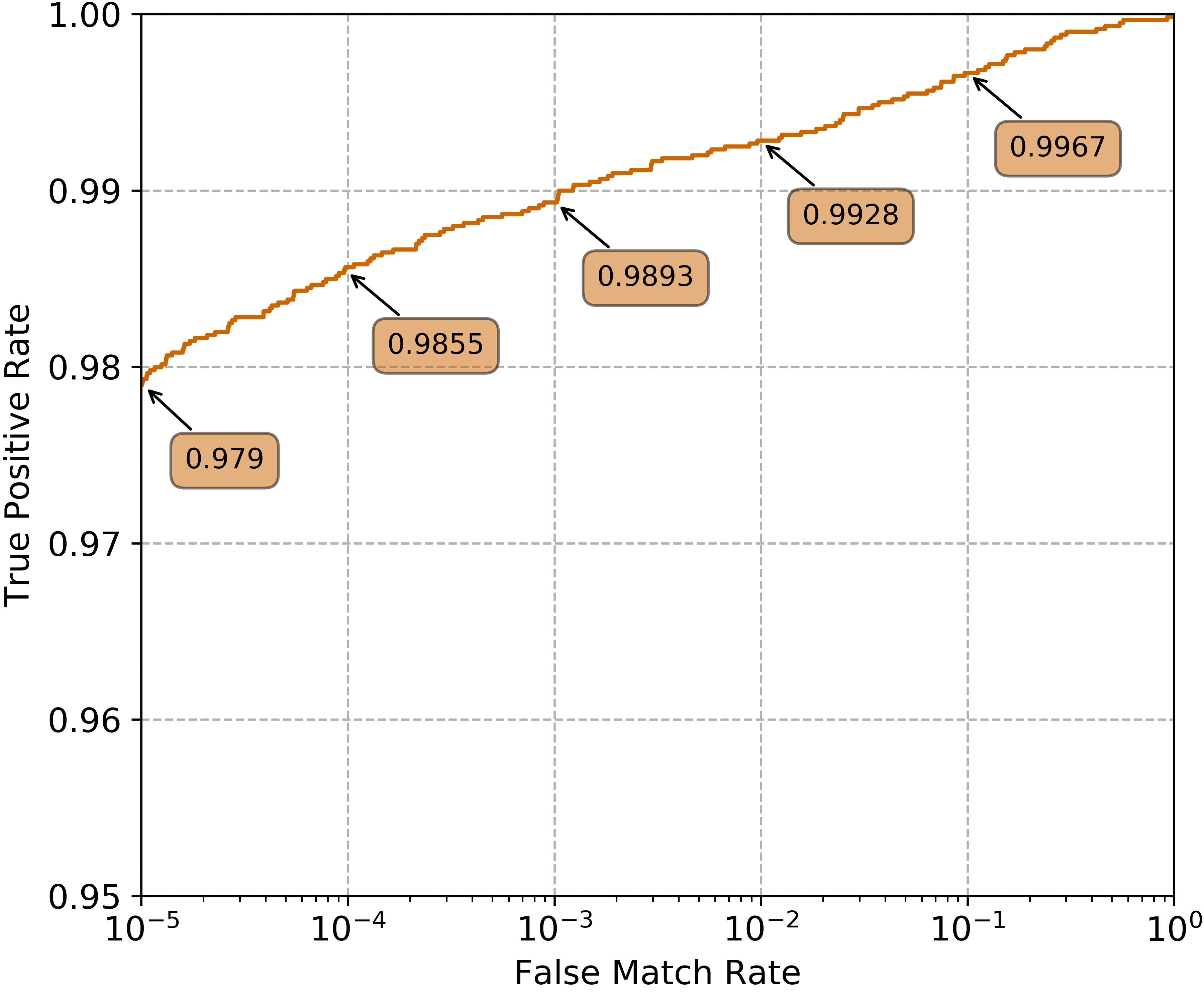}
          \caption{VGGFace2 Off-The-Shelf (Layer 27)}
      \end{subfigure}
  \end{subfigure}
  \begin{subfigure}[b]{1\textwidth}
      \begin{subfigure}[b]{0.327\textwidth}
          \centering
          \includegraphics[width=1\columnwidth]{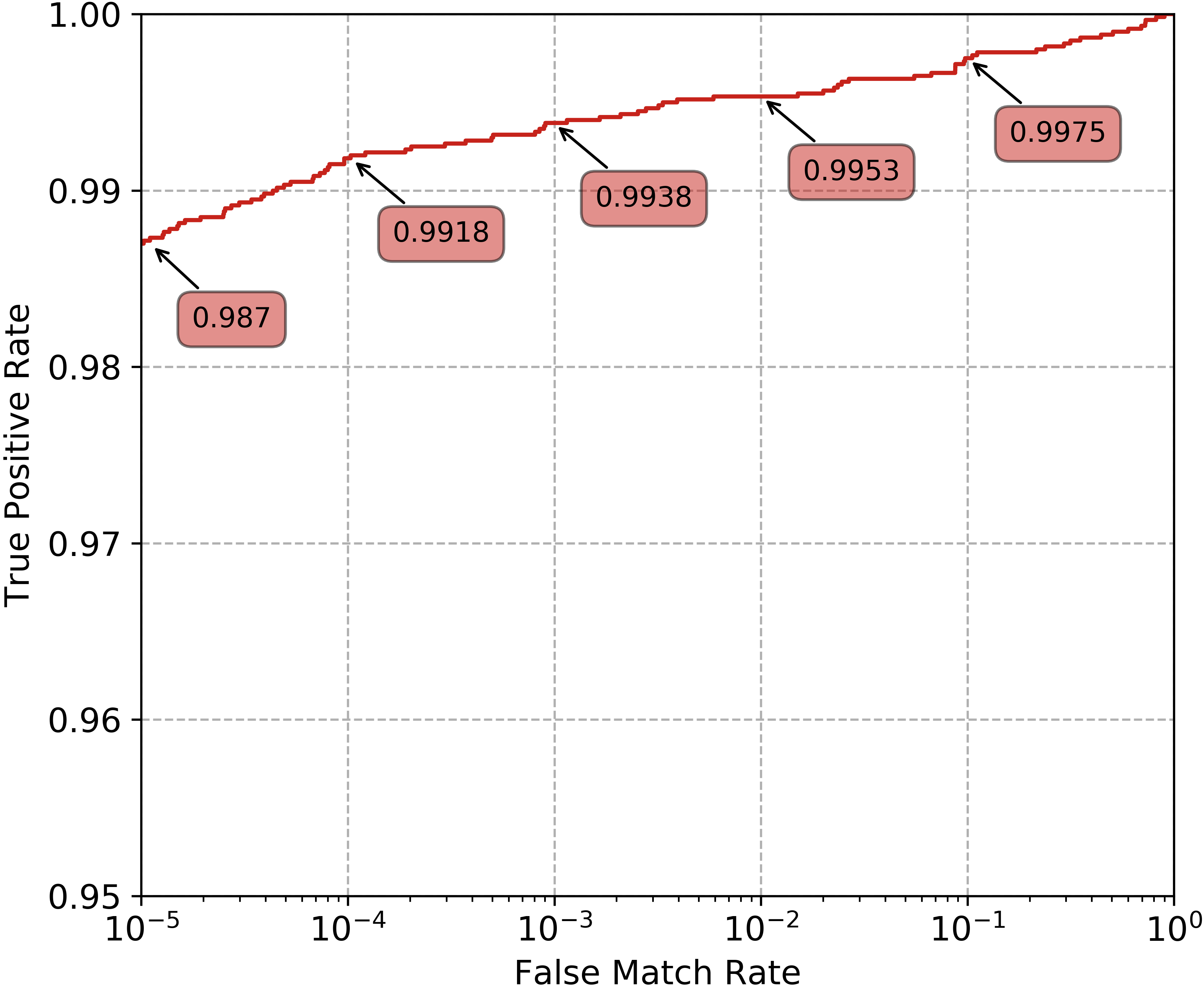}
          \caption{ImageNet Fine-tuned (Layer 23)}
      \end{subfigure}
      \hfill
      \begin{subfigure}[b]{0.327\textwidth}
          \centering
          \includegraphics[width=1\columnwidth]{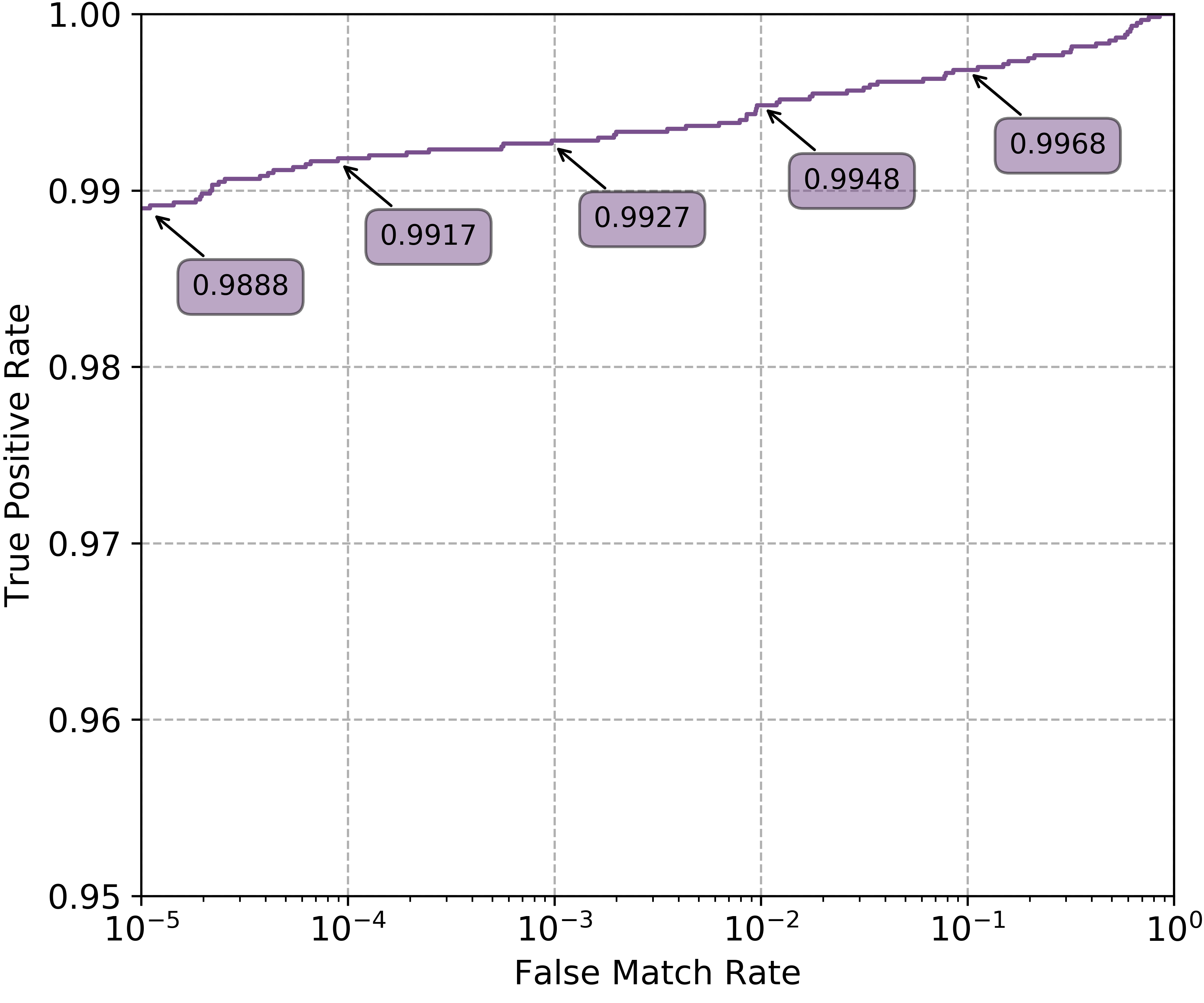}
          \caption{VGGFace2 Fine-tuned (Layer 27)}
      \end{subfigure}
      \hfill
      \begin{subfigure}[b]{0.327\textwidth}
          \centering
          \includegraphics[width=1\columnwidth]{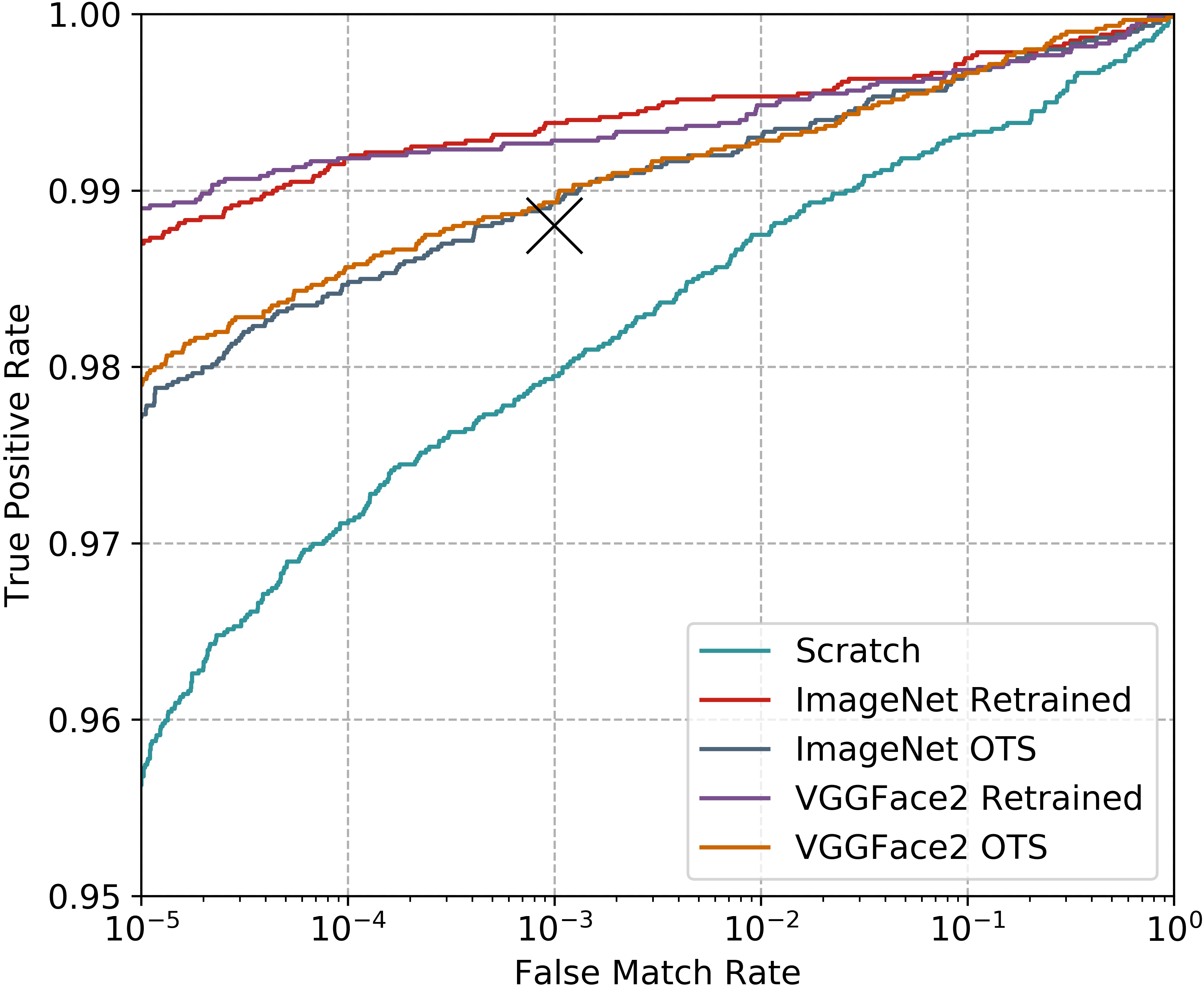}
          \caption{Combined graph of all ROC Curves}
      \end{subfigure}
      \label{fig:roc_bestlayers}
  \end{subfigure}
  \caption{ROC curves for five networks investigated in this paper. Annotated values correspond to true positive rate seen at the correspondent false match rate. Annotated by a cross in (f) is the peak recognition rate from the work by Nguyen \etal \cite{Nguyen_Access_2018} }
  \label{fig:roc}
\end{figure*}

\subsection{Comparison of results}

Although the purpose of this work is to investigate the optimal strategy to apply an example deep learning-based feature extraction (ResNet-50) for iris recognition, the obtained results can be compared to current literature to measure the performance of our approach.

In the paper by Nguyen \etal \cite{Nguyen_Access_2018}, the metric used to measure performance was the true positive rate at a false match rate of 0.1\%. To convert our results so they can be comparable to the results in \cite{Nguyen_Access_2018}, Receiver Operating Characteristic curves can be generated and the true positive rate at 0.1\% false match rate can be extracted. In their paper, the CASIA-Iris-Thousand database was used in a 70\%/30\% split in the same way as our work. We directly compare to the results obtained on this database. To do this the ROC curve for the highest performing layer for each network is created. We denote the highest performing layer as the layer that produced the highest accuracy as seen in Figure \ref{fig:results}, \ie, correct classifications/total samples in the test set. The highest performing layers are as follows: 

\begin{itemize}
	\item For the network trained from scratch, the best performing layer was \textit{layer 42}. This layer attained an accuracy of 97.03\%. The ROC Curve for this layer can be seen in Figure \ref{fig:roc}(a). Looking at this graph we see that the true positive rate at a FMR of 0.1\% ($10^{-3}$ FMR in Figure \ref{fig:roc}) is 97.93\%. 
	
	\item For the off-the-shelf ImageNet weights, the highest accuracy seen was 98.43\% using layer \textit{layer 25}. As per Figure \ref{fig:roc}(b), the true positive rate of this layer is 98.93\%. 
	
	\item The best performing layer for the off-the-shelf VGGFace2 weights saw an accuracy of 98.41\% when using layer \textit{layer 27}. This translated into a true positive rate of 98.93\% as shown in Figure \ref{fig:roc}(c).

	\item For the network that used weights that were fine-tuned from ImageNet weights, an accuracy of 99.03\% was obtained using layer \textit{layer 23}. Figure \ref{fig:roc}(d) depicts the ROC curve for this network configuration. The true positive rate for this layer is 99.38\%. 
	
	\item For the network that used weights that were fine-tuned from the VGGFace2 weights, the highest accuracy attained was 99.03\%, the same as that for the fine-tuned ImageNet network. This accuracy was achieved using layer \textit{layer 27}. As per Figure \ref{fig:roc}(e), the true positive rate of this layer is slightly lower than the fine-tuned ImageNet, at 99.27\%.
\end{itemize}

In the paper by Nguyen \etal \cite{Nguyen_Access_2018}, the highest recorded recognition rate was 98.8\% using the DenseNet architecture. In their work, they also test a shallower ResNet architecture, attaining a peak recognition rate of 98.5\%. The peak recognition rate in our experiments was using the fine-tuned ImageNet network at 99.38\%. Figure \ref{fig:roc}(f) shows all five ROC curves generated superimposed on the same graph, with the peak recognition rate seen in \cite{Nguyen_Access_2018} annotated as a black X. It can be seen from this graph that four of the five networks tested in this work perform better than the highest recorded recognition rate of \cite{Nguyen_Access_2018}. This may additionally suggest that fine-tuning of the already trained networks to the iris domain is a good approach to use deep learning-based structures in iris recognition. The use of a deeper network is also shown to be beneficial as there are more layers to extract features from, hence a higher chance of generating a better feature extractor.

\begin{figure}[h!]
\includegraphics[scale = 0.5]{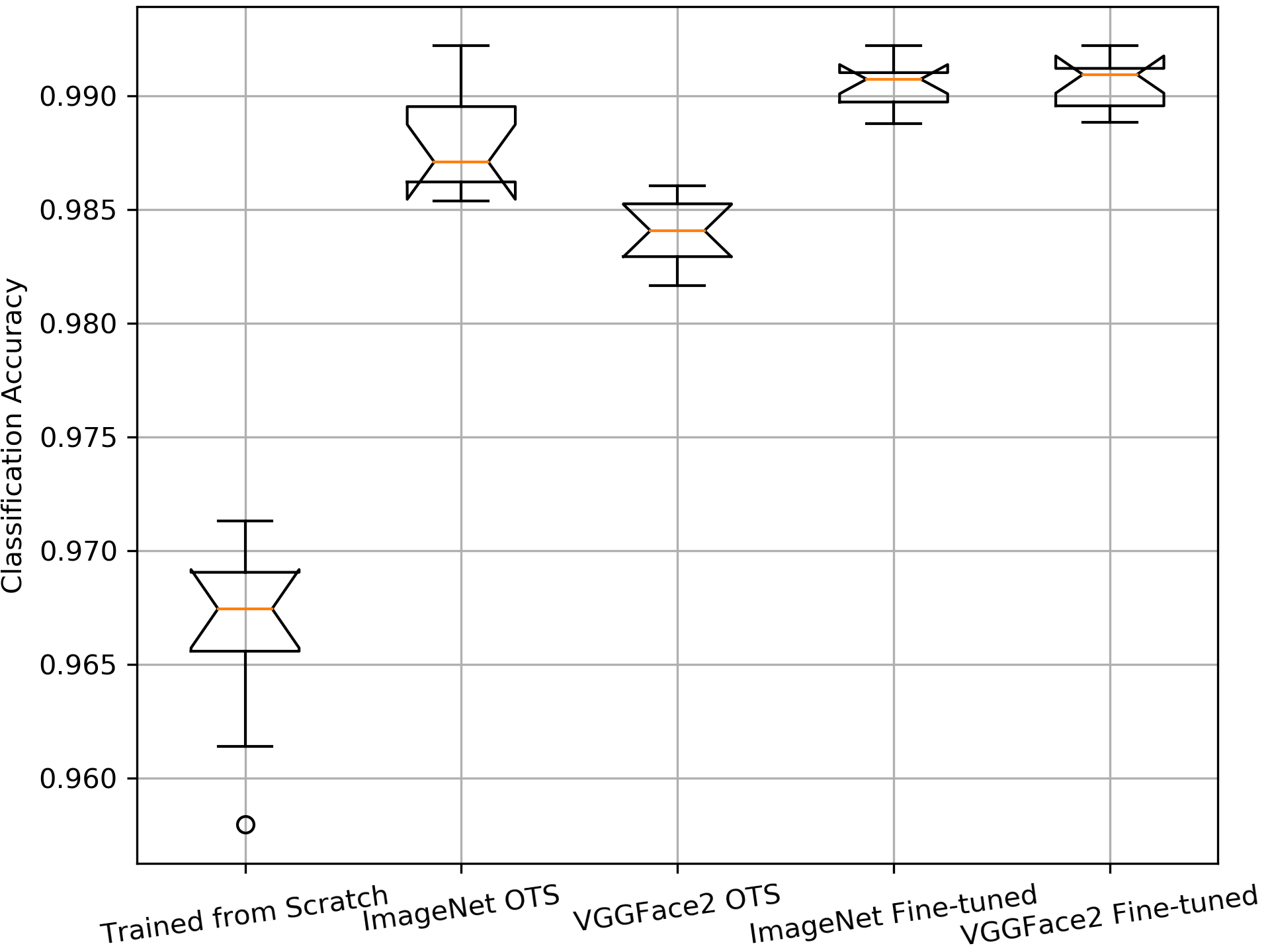}
\caption{Boxplot showing the results of 10 80\%/20\% splits of the test data for all five network configurations. }
\label{fig: boxplot}
\end{figure}

\subsection{Statistical Significance of Results}

To check the statistical significance of the results that were obtained, further analysis was done through the use of a boxplot. For the best performing layer of each network, we took the same 70\%/30\% split of the classification database and then further broke the 30\% into 10 different 80\%/20\% splits. We discarded the 20\% and then ran the classification on the 80\%. This was to check if different sub-splits of the testing data would yield similar results as what we saw on the full 30\%. The result of this experiment was Figure \ref{fig: boxplot}. 

From this we see clearly that the network trained from scratch performed the worst over all sub-splits. Figure \ref{fig: boxplot} displays something interesting though, the two off-the-shelf networks performed differently in this experiment. It can be seen that the ImageNet weights perform statistically better than the VGGFace2 weights, even though the result obtained for the full 30\% only differed by 0.02\% (ImageNet 98.43\%/VGGFace2 98.41\%). This outlines that the ImageNet weights actually perform better for this task as the results are in general slightly higher on the sub-splits. This information would not be attainable through using just the full 30\% for testing, so the creation of 10 sub-splits gives us more information about the overall performance.

For the two fine-tuned networks, the variance in performance was minimal, the upper and lower quartile range for both the fine-tuned ImageNet and VGGFace2 weights are similar and close together. The results obtained from this sub-splitting did not vary greatly. This could be due to the fine-tuning process tuning the weights to similar values. The upper quartile of the ImageNet off-the-shelf actually matches that of the two fine-tuned networks, however the range of results is larger. From this we can affirm our conclusion that the fine-tuned weights are the most performant, however, off-the-shelf weights can be employed to generate effective feature extractors also.

\section{Discussion and Conclusions}

The results presented in this paper allow us to provide the following answers to two questions posed in the introduction:

\begin{enumerate}
    \item[Q1.] It is worth using deep learning-based model trained on domain-specific images in iris recognition.
    \item[Q2.] It is better to take the best-performing model trained on either general-purpose or face images and fine-tune it to iris recognition task, rather than train own network from scratch.
\end{enumerate}

To answer these questions, we examined five different sets of weights on the popular ResNet-50 architecture and extracted features from each convolutional layer in the architecture. These sets of weights included a network trained from scratch using random weight initialization, off-the-shelf \mbox{ImageNet} weights, off-the shelf \mbox{VGGFace2} weights, fine-tuned \mbox{ImageNet} weights and fine-tuned \mbox{VGGFace2} weights. 


The reason for the observed results may be that such complex and deep structures like ResNet-50 require more samples than we had for iris recognition domain (around 360,000). And it is thus better to start with a solution for general-purpose vision problem, and then fine-tune it to the specific domain. Although this conclusion seems quite obvious, it was interesting to see that 360,000 training samples seems to be too small for training such structures from scratch.
%
%
The training dataset size clearly plays a large role in the creation of good feature extractors. Also, starting from non-domain-specific weights and fine-tuning them increases heterogeneity in training.We conclude that the weights used to classify natural scenes are a good starting point for network training as the highly variant classes used to generate these weights meant more generalized feature extractors were created, which, once fine-tuned, perform well for the task of iris recognition even in a cross-sensor scenario as presented in this paper.


In this work, we not only show the optimal training method for iris recognition, we also show that our approach is effective for iris recognition by comparing our attained results to other recent work in the area. We show that the proposed methodology shows that for four out of the five weight sets resulted in an increase in recognition rate as compared to a previous work. Although this was not the primary purpose of this paper, the improved results verifies the approach taken.

This paper follows the good practices related to reproducibility of research results. We made the performing weights publicly available for those who would like to explore the best known to us at present deep learning-based iris feature extractor \cite{aboyd-github}.

\section*{Acknowledgments}
The Titan Xp used for this research was donated by the NVIDIA Corporation. We would also like to thank V\'{i}tor Albiero for his help with this work.


{\small
\bibliographystyle{ieee}
\bibliography{refs}
}

\end{document}